\documentclass[10pt,twocolumn,letterpaper]{article}

\usepackage{cvpr}              

\usepackage{tikz}
\usepackage{algorithm,algpseudocode}
\usepackage{amsmath} 
\usepackage{bbm}
\usepackage{pifont}
\usepackage{float}
 \usepackage{multirow} 
 \usepackage{pgfplots}
\pgfplotsset{compat=1.18}
\usepackage{xcolor}
\usetikzlibrary{patterns}
\usetikzlibrary{shapes,arrows}
\usetikzlibrary{calc}
\definecolor{cvprblue}{rgb}{0.21,0.49,0.74}
\usepackage[pagebackref,breaklinks,colorlinks,allcolors=cvprblue]{hyperref}

\def\paperID{DG-EBF-8} 
\def\confName{CVPR}
\def\confYear{2026}

\title{Domain-Agnostic Feature Modulation for Semi-Supervised Domain Generalization}

\author{
Venuri Amarasinghe,
Kalinga Bandara,
Isun Randila,
Asini Jayakody\\
University of Moratuwa\\
{\tt\small \{amarasingheamvm.20, bandarammkr.20, kandegedarapmirb.20, jayakodyjaau.20\}@uom.lk}
\and
Chamuditha Jayanga Galappaththige\\
Queensland University of Technology\\
{\tt\small chamuditha.galappaththige@qut.edu.au}
\and
Ranga Rodrigo\\
University of Moratuwa\\
{\tt\small ranga@uom.lk}
}

\begin{document}
\maketitle
\begin{abstract}
Semi-supervised domain generalization (SSDG) leverages a small fraction of labeled data alongside unlabeled data to enhance model generalization. Most of the existing SSDG methods rely on pseudo-labeling (PL) for unlabeled data, often assuming access to domain labels—a privilege not always available. However, domain shifts introduce domain noise, leading to inconsistent PLs that degrade model performance. Methods derived from semi-supervised learning baselines suffer particularly from lower PL accuracy, reducing the effectiveness of unlabeled data.
We address this in a more challenging setting---domain-label agnostic SSDG---,where domain labels for unlabeled data are not available during training. First, we propose a feature modulation strategy that enhances class-discriminative features while suppressing domain-specific information. This modulation shifts features toward Similar Average Representations---a modified version of class prototypes---that are robust across domains, encouraging the classifier to distinguish between closely related classes and feature extractor to form tightly clustered, domain-invariant representations. Second, to mitigate domain noise and improve pseudo-label accuracy, we introduce a loss-scaling function that dynamically lowers the fixed confidence threshold for pseudo-labels, optimizing the use of unlabeled data. With these key innovations, our approach achieves significant improvements on four major domain generalization benchmarks—even without domain labels. We will make the code available.
\end{abstract}    
\section{Introduction}
\label{sec:intro}

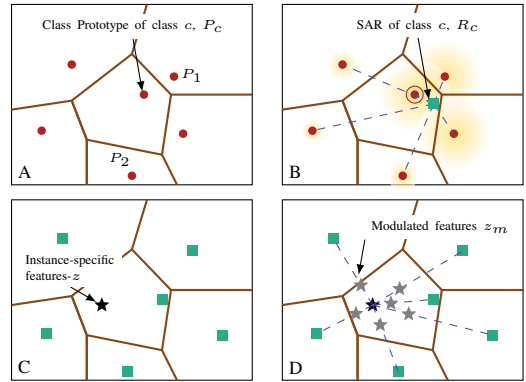
\begin{figure}[ht]
     \centering
     \scriptsize
\begin{tikzpicture}[scale=0.4]
\tikzset{every node/.style={font=\scriptsize, execute at begin node=\setlength{\baselineskip}{0.9em}}}
\definecolor{br}{rgb}{0.7, 0.13, 0.13}	
\definecolor{sb}{RGB}{63, 81, 181}  
\definecolor{dl}{RGB}{255, 214, 73}  
\definecolor{brown}{RGB}{139, 69, 19} 
\definecolor{jg}{RGB}{41, 171, 135}

\begin{scope}    
	\coordinate (o) at (-3.5,-2.5);
	\draw (o) node[anchor=south west] {A}  rectangle ++(8,6);
	\coordinate (A) at (-1.5, 1.5);
	\coordinate (B) at (1.9, 1.1);
	\coordinate (C) at (-2.5, -0.7);
	\coordinate (D) at (0.9, 0.5);
	\coordinate (E) at (2.2, -0.8);
	\coordinate (F) at (0.5, -2.2);
	
	\draw [thick, brown] (-3.5,0) -- (-1.5, 0.3) coordinate (p3)-- (0.5,1.85) coordinate (p4)  -- (1,3.5);
	\draw [thick, brown] (p4)  -- (1.8,0.5)  coordinate (p5) -- (4.5,0.5);
	\draw [thick, brown] (p5)  --  (1.5,-1.5)  coordinate (p1) --  (2,-2.5);
	\draw [thick, brown] (p3) -- (-1,-1) coordinate (p2)  --  (-1,-2.5);	
	\draw [thick, brown] (p1) -- (p2);
	\draw [thick, brown] (p3)-- (-1,-1) --  (-1,-2.5);
	
	\foreach \coord in {A, B, C, D, E, F} {
	    \fill[br] (\coord) circle (4pt); 
	}
	\draw[latex-] (D) ++(120:0.1) -- ++(100:1.8) node[anchor=south, text width=2.5cm, align=center, fill = white] {\tiny Class Prototype of class $c$, $P_c$};

        \node[anchor=south, text width=0.3cm, align=center] at (2.5, 0.7){\tiny $P_1$};
        \node[anchor=south, text width=0.3cm, align=center] at (0.1, -2.1){\tiny $P_2$};
	    
\end{scope}

\begin{scope}[shift={(9, 0)}]
	\coordinate (o) at (-3.5,-2.5);
	\draw (o) node[anchor=south west] {B} rectangle ++(8,6);
	\coordinate (A) at (-1.5, 1.5);
	\coordinate (B) at (1.9, 1.1);
	\coordinate (C) at (-2.5, -0.7);
	\coordinate (D) at (0.9, 0.5);
	\coordinate (E) at (2.2, -0.8);
	\coordinate (F) at (0.5, -2.2);
        \coordinate (avg) at ($(B)!0.70!(D)!0.33!(E)$);

        \coordinate (G) at (0.9, 0.5);

        \shade[ball color=dl!80, opacity=0.6,  shading=radial, inner color=dl, outer color=white]  (B) ellipse (1.2cm and 1.2cm);
	\shade[ball color=dl!80, opacity=0.6,  shading=radial, inner color=dl, outer color=white]  (D) ellipse (1.2cm and 1.2cm);	
	\shade[ball color=dl!80, opacity=0.6,  shading=radial, inner color=dl, outer color=white]  (E) ellipse (1.2cm and 1.2cm);
        \shade[ball color=dl!80, opacity=0.6,  shading=radial, inner color=dl, outer color=white]  (A) ellipse (0.5cm and 0.5cm);
	\shade[ball color=dl!80, opacity=0.6,  shading=radial, inner color=dl, outer color=white]  (C) ellipse (0.4cm and 0.4cm);	
	\shade[ball color=dl!80, opacity=0.6,  shading=radial, inner color=dl, outer color=white]  (F) ellipse (0.5cm and 0.5cm);

	\draw [thick, brown] (-3.5,0) -- (-1.5, 0.3) coordinate (p3)-- (0.5,1.85) coordinate (p4)  -- (1,3.5);
	\draw [thick, brown] (p4)  -- (1.8,0.5)  coordinate (p5) -- (4.5,0.5);
	\draw [thick, brown] (p5)  --  (1.5,-1.5)  coordinate (p1) --  (2,-2.5);
	\draw [thick, brown] (p3) -- (-1,-1) coordinate (p2)  --  (-1,-2.5);	
	\draw [thick, brown] (p1) -- (p2);
	\draw [thick, brown] (p3)-- (-1,-1) --  (-1,-2.5);
	
	\foreach \coord in {B, D, E, A, C, F} {
	    \fill[br] (\coord) circle (4pt); 
	    \draw[dashed, sb] (avg) -- (\coord);
	}
        \draw[br] (G) circle (8pt);

	\foreach \coord in {avg} {
	    \fill[jg] (\coord) ++(-5pt, -5pt) rectangle ++(10pt, 10pt);
	}
	\draw[latex-] (avg) ++(140:0.2) -- ++(100:2.1) node[anchor=south, text width=2.0cm, align=center, fill=white] {\tiny SAR of class $c$, $R_c$};		
\end{scope}

\begin{scope}[shift={(0, -6.5)}]
	\coordinate (o) at (-3.5,-2.5);
	\draw (o)   node[anchor=south west] {C} rectangle ++(8,6);
	\coordinate (A) at (-1.5, 1.5);
	\coordinate (B) at (1.9, 1.1);
	\coordinate (C) at (-2.5, -0.7);
	\coordinate (D) at (0.9, 0.5);
	\coordinate (E) at (2.2, -0.8);
	\coordinate (F) at (0.5, -2.2);
        \coordinate (avg) at ($(B)!0.70!(D)!0.33!(E)$);

	\coordinate (P) at (-1.8, 2.2);
	\coordinate (Q) at (2.5, 1.8);
	\coordinate (R) at (-2.35, -0.95);
	\coordinate (T) at (3.5, -1);
	\coordinate (U) at (0.35, -2.2);	
	
	\coordinate (data) at (-0.5,0);

	\draw [thick, brown] (-3.5,0) -- (-1.5, 0.3) coordinate (p3)-- (0.5,1.85) coordinate (p4)  -- (1,3.5);
	\draw [thick, brown] (p4)  -- (1.8,0.5)  coordinate (p5) -- (4.5,0.5);
	\draw [thick, brown] (p5)  --  (1.5,-1.5)  coordinate (p1) --  (2,-2.5);
	\draw [thick, brown] (p3) -- (-1,-1) coordinate (p2)  --  (-1,-2.5);	
	\draw [thick, brown] (p1) -- (p2);
	\draw [thick, brown] (p3)-- (-1,-1) --  (-1,-2.5);
	
	\foreach \coord in {P, Q, R, T, U, avg} {
	    \fill[jg] (\coord) ++(-5pt, -5pt) rectangle ++(10pt, 10pt);
	}
	
	\draw (data) node[star, star points=5, star point ratio=2.25, fill=black, minimum size=6pt, inner sep=0] {};
	\draw[latex-] (data) ++(150:0.2) -- ++(150:1) node[anchor=south, text width=1.2cm, align=left,fill=white] {\tiny Instance-specific features-$z$};

\end{scope}

\begin{scope}[shift={(9, -6.5)}]
	\coordinate (o) at (-3.5,-2.5);
	\draw (o) node[anchor=south west] {D} rectangle ++(8,6);
	\coordinate (A) at (-1.5, 1.5);
	\coordinate (B) at (1.9, 1.1);
	\coordinate (C) at (-2.5, -0.7);
	\coordinate (D) at (0.9, 0.5);
	\coordinate (E) at (2.2, -0.8);
	\coordinate (F) at (0.5, -2.2);
        \coordinate (avg) at ($(B)!0.70!(D)!0.33!(E)$);

	\coordinate (P) at (-1.8, 2.2);
	\coordinate (Q) at (2.5, 1.8);
	\coordinate (R) at (-2.35, -0.95);
	\coordinate (T) at (3.5, -1);
	\coordinate (U) at (0.35, -2.2);	
	
	\coordinate (data) at (-0.5,0);

	\draw [thick, brown] (-3.5,0) -- (-1.5, 0.3) coordinate (p3)-- (0.5,1.85) coordinate (p4)  -- (1,3.5);
	\draw [thick, brown] (p4)  -- (1.8,0.5)  coordinate (p5) -- (4.5,0.5);
	\draw [thick, brown] (p5)  --  (1.5,-1.5)  coordinate (p1) --  (2,-2.5);
	\draw [thick, brown] (p3) -- (-1,-1) coordinate (p2)  --  (-1,-2.5);	
	\draw [thick, brown] (p1) -- (p2);
	\draw [thick, brown] (p3)-- (-1,-1) --  (-1,-2.5);
	
	\draw (data) node[star, star points=5, star point ratio=2.25, fill=black, minimum size=6pt, inner sep=0] {};
	
	\foreach \coord in {P, Q, R, T, U, avg} {
	    \fill[jg] (\coord) ++(-5pt, -5pt) rectangle ++(10pt, 10pt);
	    	    \draw[dashed, sb] (data) -- (\coord) node[pos=0.3, star, star points=5, star point ratio=2.25, fill=gray, minimum size=6pt, inner sep=0] {};
	}
	
	\draw[latex-]  (-1,0.8) ++(80:0.2) -- ++(80:1.2) node[anchor=south west, text width=1.9cm, align=left, fill=white] {\tiny Modulated features $z_m$};		
	
\end{scope}

\end{tikzpicture}
      \caption{Feature modulation towards SARs in the classwise separated feature space. Polygon lines denote class boundaries.  
      {\bf A}: Class prototypes $P_c$ in the feature space.
      {\bf B}: SAR of class C (Sec.~\ref{Similar Average Representations}) generated by averaging class prototypes based on similarity. The glow intensity represents the similarity.
      {\bf C}: Instance-specific features $z$ in the feature space with SARs of all classes.
      {\bf D}: Feature modulation shifts $z$ more towards its true class.}
      \vspace{-0.5cm}
     \label{fig:intrduction_image}
 \end{figure}

Domain shift—the discrepancy between the distributions of training and testing data, violating the independent and identically distributed (i.i.d.) assumption—is a fundamental challenge in deep learning models \cite{dosovitskiy2020image, he2016deep}. Domain Generalization (DG) \cite{li2017deeper, Zhang_2021_CVPR, gulrajani2020search, xu2021fourier} and Domain Adaptation \cite{zhang2013domain, ganin2016domain, motiian2017few} have emerged as two key paradigms to address this issue. DG aims to develop models that generalize to unseen domains using multiple labeled source domains, whereas domain adaptation leverages knowledge from both source and target domains to improve test performance.

To enhance generalization across domains, DG methods typically focus on learning domain-invariant representations using techniques such as invariant risk minimization \cite{arjovsky2019invariant}, meta-learning \cite{dou2019domain, li2019episodic, shu2021open}, data augmentation \cite{xu2021fourier,volpi2018generalizing, zhong2022adversarial, wang2021augmax}, and adversarial feature learning \cite{volpi2018generalizing, li2018domain, li2018deep}. However, these approaches rely heavily on labeled data from multiple source domains, which is often scarce in real-world settings.
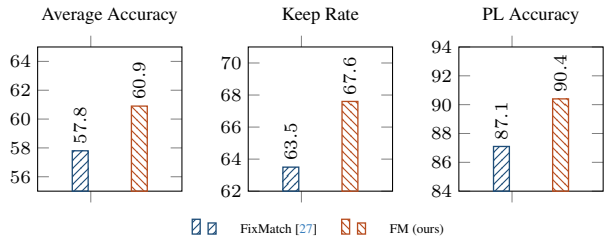
\begin{figure}[h]
     \centering

\definecolor{fixmatchFill}{HTML}{A3C1E9}  
\definecolor{fixmatchLine}{HTML}{1F4E79}  
\definecolor{dgwmFill}{HTML}{C4DFAA}      
\definecolor{dgwmLine}{HTML}{4A7023}      
\definecolor{oursFill}{HTML}{F9C5A5}      
\definecolor{oursLine}{HTML}{B84E20}      
\definecolor{dgwmoursFill}{HTML}{D6BBE8}      
\definecolor{dgwmoursLine}{HTML}{5A2C8A}      

\scriptsize
\begin{tikzpicture}
    \begin{axis}[
        width=3.5cm,
        height=3.5cm,
        ybar,
        bar width=6pt,
        ylabel={},
        ymin=55, ymax=65,
        symbolic x coords={1, 2},    
        xtick=data,
        nodes near coords,
        every node near coord/.append style={rotate=90, anchor=west},
        title={Average Accuracy},
        enlarge x limits=1.4,   
        xticklabel=\empty, 
    ]
        \addplot[fill=fixmatchFill, draw=fixmatchLine, pattern=north east lines, pattern color=fixmatchLine, line width=0.5pt] coordinates {(1, 57.8)};
        \addplot[fill=oursFill, draw=oursLine, pattern=north west lines, pattern color=oursLine, line width=0.5pt] coordinates {(2, 61.0)};
                     \end{axis}
    \begin{axis}[
        at={(5.6cm,0)}, 
        width=3.5cm,
        height=3.5cm,
        ybar,
        bar width=6pt,
        ymin=84, ymax=94,
        symbolic x coords={1, 2},
        xtick=data,
        nodes near coords,
        every node near coord/.append style={rotate=90, anchor=west},        
        title={PL Accuracy},
        enlarge x limits=1.4, 
        xticklabel=\empty, 
    ]
        \addplot[fill=fixmatchFill, draw=fixmatchLine, pattern=north east lines, pattern color=fixmatchLine, line width=0.5pt] coordinates {(1, 87.1)};
        \addplot[fill=oursFill, draw=oursLine, pattern=north west lines, pattern color=oursLine, line width=0.5pt] coordinates {(2, 90.4)};
    \end{axis}

    \begin{axis}[
        at={(2.8cm,0)}, 
        width=3.5cm,
        height=3.5cm,
        ybar,
        bar width=6pt,
        ymin=62, ymax=71,
        symbolic x coords={1, 2}, 
        xtick=data,
        nodes near coords,
        every node near coord/.append style={rotate=90, anchor=west},        
        title={Keep Rate},
        enlarge x limits=0.5,
        legend style={
            at={(0.5,-0.15)}, 
            anchor=north, 
            legend columns=4, 
            /tikz/every node/.append style={align=center, font=\tiny}, 
            column sep=1em, 
            draw=none,
        },   
        enlarge x limits=1.4,   
        xticklabel=\empty, 
    ]
        \addplot[fill=fixmatchFill, draw=fixmatchLine, pattern=north east lines, pattern color=fixmatchLine, line width=0.5pt] coordinates {(1, 63.5)};
        \addplot[fill=oursFill, draw=oursLine, pattern=north west lines, pattern color=oursLine, line width=0.5pt] coordinates {(2, 67.6)};
        \legend{FixMatch~\cite{sohn2020fixmatch},  FM (ours)}    
    \end{axis}

    
\end{tikzpicture}

      \vspace{-0.2cm}
      \caption{Average accuracy, keep rate, and pseudo-label (PL) accuracy. We achieve higher average accuracy as a result of the increased PL accuracy and higher keep rate. Note that different $y$-axis ranges were chosen to enhance the visualization. }
      \vspace{-0.5cm}
     \label{fig:bar_graph}
 \end{figure}
Semi-supervised learning (SSL) presents a viable solution to address labeled data scarcity by leveraging abundant unlabeled data alongside a limited set of labeled instances \cite{sohn2020fixmatch, berthelot2019mixmatch, tarvainen2017mean}. Recent SSL methods \cite{sohn2020fixmatch, wang2022freematch, zhou2023hypermatch, Chen_2023_CVPR} predominantly utilize pseudo-labeling and consistency regularization. Pseudo-labeling \cite{lee2013pseudo} assigns artificial labels to unlabeled data with high prediction confidence, while consistency regularization ensures that perturbed versions of the same data receive consistent predictions. Although DG methods struggle when a majority of training data is unlabeled, SSL methods naturally exploit unlabeled data. However, their effectiveness is hindered by noisy pseudo-labels and the inability to extract robust domain-invariant features, making them inferior to fully supervised DG methods.

Semi-supervised domain generalization (SSDG) \cite{Galappaththige_2024_CVPR, zhou2023semi, zhang2024semi, wang2023better} aims to bridge this gap by combining DG and SSL to improve generalization to unseen domains while mitigating the domain shift, poor data utilization, and pseudo-label noise inherent in SSL. Recent SSDG methods introduce multi-view consistency learning, stochastic classifiers \cite{zhou2023semi}, feature-based conformity, and semantic alignment constraints \cite{Galappaththige_2024_CVPR}. However, these methods rely on domain labels, which are difficult to obtain in practice. Moreover, their utilization of unlabeled data is suboptimal in early training stages, as the number of valid pseudo-labels passing the fixed confidence threshold is low. Simply lowering this threshold to incorporate more unlabeled data results in increased pseudo-label noise and inaccurate predictions which result in inferior DG performances. This presents a fundamental \textit{dilemma}—balancing accurate pseudo-labels with higher unlabeled data utilization.

To address this challenge, we propose a novel SSDG method 
that enhances pseudo-label accuracy and improves unlabeled data utilization under a more challenging setting where domain labels are unavailable during training. Our approach is motivated by the intuition that class representations in each domain comprise two signals: a class-specific signal, and a domain-related signal. We aim to isolate and emphasize the class-specific signal while minimizing domain information propagation to the classifier through a proposed \textbf{feature modulator}. 

We also notice that some features are common to more than one class. For example, dogs and cats have many similar features. Our method encourages the classifier to focus on class-specific features while minimizing dependence on features common to multiple classes. We adopt similar average representations SAR---a modified version of class prototypes---to reduce the effect of domain-specific features, ensuring the classifier focuses on class-specific information and improves domain generalization. The feature extractor and the classifier are trained while shifting extracted features toward SARs using the feature modulator (Fig.~\ref{fig:intrduction_image}). This approach helps the classifier distinguish similar classes more accurately while guiding the feature extractor to form clearer class boundaries. As a result, it improves pseudo-label accuracy by \textbf{3.3\%} over the baseline (Fig.~\ref{fig:bar_graph}).


The intervention of the feature modulator for better pseudo-labels allows us to incorporate more unlabeled data for model training when coupled with our proposed loss scaling function that dynamically adjusts the effect of pseudo-labels. Instead of naively lowering the threshold—introducing label noise—we scale loss gradients based on pseudo-label confidence and uncertainty scores. This strategy increases the portion of unlabeled data used for training by \textbf{4.1\%} compared to the baseline (Fig.~\ref{fig:bar_graph}). The improved data utilization with accurate pseudo labels improves the DG capability of the learned model. In summary, our contributions are:
\begin{itemize}
    \item[--] We propose a feature modulator that enhances class-specific feature representations, leading to more accurate pseudo-labels without relying on domain labels.
    \item[--] We introduce a loss scaling function that dynamically lowers the effect 
    of low confidence pseudo-labels while reducing label noise, improving unlabeled data utilization.
    \item[--] We address the SSDG problem under more challenging setting where domain labels are unavailable and demonstrate the effectiveness of our method on four major domain generalization benchmarks.
    
\end{itemize}
\section{Related Work}
\label{sec:formatting}
 
\noindent \textbf{Domain Generalization:} With the objective of creating models that can generalize across unseen domains, the DG methods can be subdivided into data augmentation \cite{xu2021fourier,volpi2018generalizing}, feature augmentation \cite{zhou2024mixstyle} and meta-learning \cite{dou2019domain, li2019episodic, shu2021open} based approaches. Data augmentation \cite{xu2021fourier,volpi2018generalizing} based DG models generate diverse training samples to improve cross-domain generalization. Mixing instance-level feature statistics in \cite{zhou2024mixstyle}, can be identified as a feature augmentation method for DG. The primary objective of these data and feature augmentation methods is to learn domain-invariant features. Subsequent works apply meta-learning to DG by designing tasks and losses to generalize across domains \cite{dou2019domain,li2019episodic,shu2021open}. However, the performance of these fully supervised DG methods relies on the fully labeled data in multiple source domains, limiting their effectiveness in practical semi-supervised settings.\\
\noindent \textbf{Semi-Supervised Learning:} Semi-Supervised Learning (SSL) addresses the problem of scarcity of labeled data by using a large amount of unlabeled data with a few labeled data. In recent literature, consistency regularization \cite{abuduweili2021adaptive,lee2022contrastive}, entropy minimization~\cite{grandvalet2004semi} and pseudo-labeling-based approaches show the most promising results in SSL. Consistency regularization forces the model to give the same prediction to different perturbed versions \cite{berthelot2019mixmatch,sohn2020fixmatch,xie2020unsupervised} of the same input. 
Pseudo-labeling assigns labels to unlabeled data, selecting only high-confidence predictions. MixMatch \cite{berthelot2019mixmatch} unifies existing data augmentations, pseudo-labeling, and mixup \cite{zhang2018mixup} to achieve both consistency regularization and entropy minimization. RemixMatch \cite{berthelot2019remixmatch} shows that introducing strong augmentations boosts the performance. FixMatch \cite{sohn2020fixmatch} is a strong SSL baseline for SSDG which combines pseudo-labeling and consistency regularization. FreeMatch \cite{wang2022freematch} extends FixMatch \cite{sohn2020fixmatch} showing that adaptive thresholds perform better than fixed thresholds for pseudo-labeling. FlexMatch \cite{zhang2021flexmatch} improves FixMatch to utilize more unlabeled data by introducing curriculum pseudo labeling, while FullMatch \cite{chen2023boosting} uses entropy meaning loss (EML) and adaptive negative learning (ANL) for a better leveraging of unlabeled data. HyperMatch \cite{zhou2023hypermatch} separates pseudo-labels into noisy and clean pseudo-labels and leverages both of them to learn better-clustered feature representations. However, efforts to improve pseudo-label quality often ignore the number of low-confidence samples discarded, leading to poor use of unlabeled data \cite{sohn2020fixmatch,wang2022freematch,zhou2023hypermatch}. Conversely, some SSL methods \cite{zhang2021flexmatch,chen2023boosting}  prioritize data utilization but compromise pseudo-label accuracy. \\
\noindent \textbf{Semi-Supervised Domain Generalization:} SSL and semi-supervised domain generalization (SSDG) leverage unlabeled data during model training while SSDG faces a significant challenge because the data comes from different source domains. SSDG is a relatively under-explored research area, yet it presents a more realistic real-world scenario. StyleMatch \cite{zhou2023semi} extends FixMatch \cite{sohn2020fixmatch} for SSDG by incorporating stochastic modeling to reduce over-fitting in scarce labels and multi-view consistency learning to enhance DG.
The feature-based conformity and semantic alignment loss functions introduced in \cite{Galappaththige_2024_CVPR} improve pseudo-label accuracy, leveraging the domain labels as well. MultiMatch \cite{qi2024multimatch} further extends FixMatch \cite{sohn2020fixmatch} to a multi-task learning framework, enhancing the pseudo-label quality by treating individual domains as separate tasks and introducing a global task to generalize to unseen domains. 
In \cite{zhang2024semi}, the authors study SSDG for a new paradigm, known and unknown classes, while authors in \cite{wang2023better} address a different SSDG setting where only one source domain is fully labeled. Their approach involves a joint domain-aware label and dual-classifier system, which improves pseudo-labels by using a separate classifier for pseudo-labeling and combining its output probabilities with similarity to its class representation from a domain-aware class representation bank. The authors in \cite{Galappaththige_2025_WACV} propose a method  to generate accurate pseudo-labels by domain-level information vectors and utilize them to learn a soft domain-aware mask to modulate the domain-shared classifier weights. However, the methods in \cite{Galappaththige_2024_CVPR,Galappaththige_2025_WACV} require domain labels to generate domain-aware class prototypes and domain-level information vectors respectively, imposing an additional constraint on the unlabeled data. Existing works train the model using multiple domain data, where the classifier learns some domain information of source domains as well. In our work, we introduce our novel method, feature modulation which improves generalization by de-emphasizing the features containing domain information while keeping the class-specific information.

\section{Methodology}

\begin{figure*}[h]
     \centering
     \includegraphics[width=\linewidth]{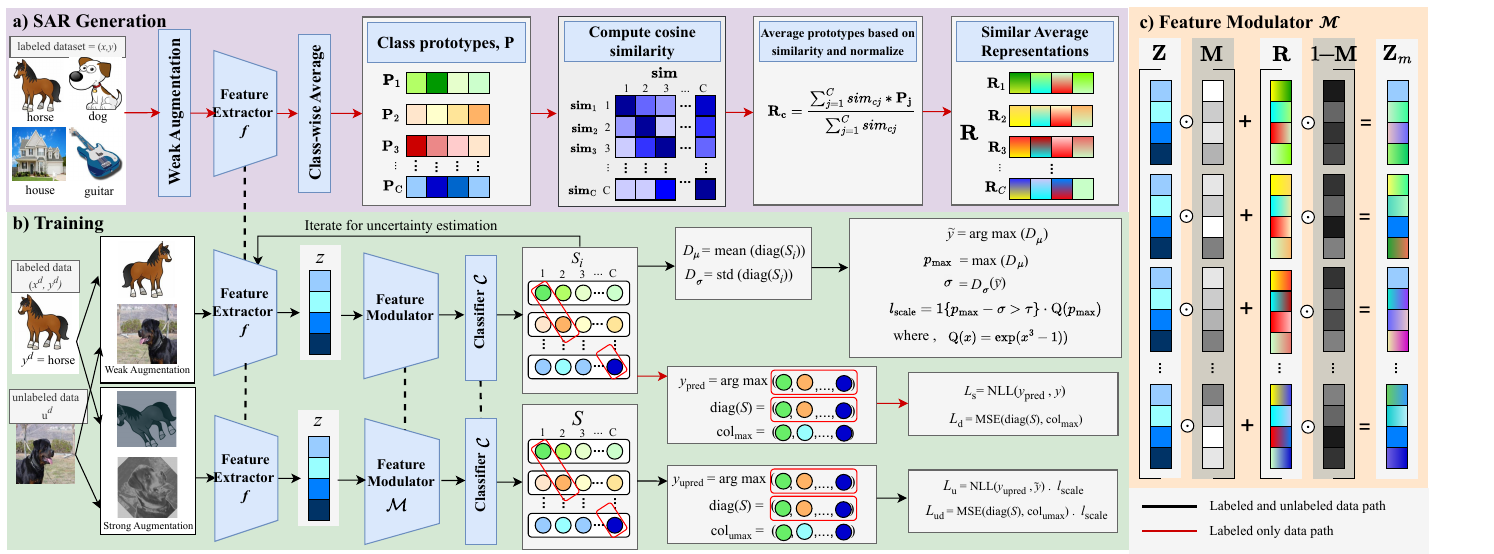}
      \caption{Proposed Architecture: a) Similar Average Representation (SAR (Sec.~\ref{Similar Average Representations})) generation. b) Complete model during training. c) Feature Modulator (Sec.~\ref{Feature Modulation}) combines instance-specific features ($z$) and SAR ($\mathbf{R}$) to soft-eliminate domain noise using our modulation strategy. This results in modulated features $\mathbf{Z}_m$. We input this $\mathbf{Z}_m$ to the shared classifier $\mathcal{C}$ and get the probabilities for each class. With these probabilities, we compute the MSE and NLL losses for both labeled and unlabeled data. For unsupervised losses, we consider loss scaling (Sec.~\ref{uncertanity_and_loss_scale})  to maximize unlabeled data utilization. Note that our method is agnostic of domain labels. }
       \vspace{-0.3cm}
     \label{fig:method}
 \end{figure*}

\noindent \textbf{Problem Setting: } We adopt notation from \cite{Galappaththige_2024_CVPR}. Let $\mathcal{X}$ and $\mathcal{Y}$  denote the input and label space, respectively, where $\mathrm {y} \in \mathcal{ Y}= \{1,2, \dots, C\}$ is the class label. A joint distribution $\mathcal{P(X,Y)}$ over $\mathcal{X} \times\mathcal{Y}$ is defined as a domain $d$. We only consider the distribution shift in $\mathcal{P(X)}$ while $\mathcal{P(Y)}$ remains the same. Following standard SSDG protocol~\cite{Galappaththige_2024_CVPR,Galappaththige_2025_WACV, zhou2023semi}, we use multiple source domains during training and an unseen domain as the target. We denote $d^L = \{(x^d_i, y^d_i)\}_{i=1}^{n_L}$ as the labeled part and $d^U = \{(u^d_i)\}_{i=1}^{n_U}$ as the unlabeled part of data for each source domain $ d \in \{d_1, d_2, \dots, d_{tr}\}$, where $x^d_i,u^d_i \in R^{(3\times H \times W)}$ are labeled and unlabeled input images. The unlabeled part contains more data than the labeled part i.e., $\mathrm n_U \gg n_L$. Our goal is to use the training (source) domains to learn a domain-generalizable model $F$ that can accurately predict on out-of-distribution data, whose examples are drawn from $\mathcal{P(X}^{d_{te}},\mathcal {Y}^{d_{te}})$, where $d_{te}$ denotes the target domain. Our model  $ F= \mathcal C ( \mathcal{M} ( f))$ comprises of feature extractor $f$, feature modulator $\mathcal{M}$, and classifier $\mathcal C$. 

\noindent\textbf{Our Motivation:} We observe that the feature spaces of certain classes can be closely aligned due to shared domain characteristics or overlapping class features (see Fig.~\ref{fig:intrduction_image}, A). Some class prototypes, such as \(\mathbf{P_1} \) and \( \mathbf {P_c}\), are closer to each other than to other classes, making classification more challenging. A higher degree of class separation in the feature space leads to a more accurate classification. Our goal is to enhance generalization by: 1) training the feature extractor to effectively separate classes, regardless of domain differences, and  2) training the classifier to focus on unique class-specific features while disregarding domain-related and shared class information.

To achieve this, we propose feature modulation, a strategy that blends features from similar classes with the originally extracted features, exposing the classifier to more challenging classification scenarios where class boundaries become less distinct. For instance, cats and dogs share several visual features, causing their feature spaces to be closely aligned, which can lead to misclassification.  To address this, we introduce a modulation process where a portion of the features from one class (e.g., a cat) is incorporated into another class (e.g., a dog) while maintaining the original class label. This modulation is guided by a learnable matrix and the cosine similarity between class prototypes. By shifting features toward their SARs (Sec.~\ref{Similar Average Representations}), the classifier learns to focus on distinct class characteristics while disregarding shared features and domain information.  Training on these feature-modulated samples forces the classifier to become more robust in distinguishing between visually similar classes, ultimately improving classification performance and generalization.  

\subsection{Feature Modulation for SSDG}
An overview of our method is shown in Fig.~\ref{fig:method}.
Our approach consists of four key components: Similar Average Representations (SAR) \( \mathbf{R} \) (see Fig.~\ref{fig:method}.a), the feature modulator \( \mathcal{M} \) (see Fig.~\ref{fig:method}.c), uncertainty estimation of pseudo-labels, and the unlabeled loss scaling function \( l_{\mathrm{scale}} \) (see Fig.~\ref{fig:method}).  First, we compute SAR (shown by the green square in Fig.\ref{fig:intrduction_image}.B) for each class using features extracted from labeled data, as described in Sec.~\ref{Similar Average Representations}. The feature modulator \( \mathcal{M} \) serves as an intermediary between the feature extractor and the classifier (see Fig.\ref{fig:method}.b), reducing the propagation of domain-specific information from the feature extractor to the classifier. The modulating matrix $\mathbf{M}$ is a learnable (i.e., dynamically updating) matrix that updates when minimizing the training loss. The modulation process adjusts the extracted instance-specific features \( z \) to enhance class-discriminative characteristics while suppressing domain-specific variations (modulated features ${z_m}$ are shown by the gray starts in Fig.\ref{fig:intrduction_image}, D). This is achieved using SAR along with a learnable feature modulation matrix \( \mathbf{M} \), which guides the transformation to ensure improved generalization across domains.

We then compute pseudo-labels for unlabeled data using the modulated features of weakly augmented unlabeled samples, as described in Sec.~\ref{Pseudo labeling}. With improved pseudo-label accuracy resulting from reduced domain interference, we employ the loss scaling function \( l_{\mathrm{scale}} \) Eq.~\eqref{eq:lscale} to leverage less confident pseudo-labels.  To further refine pseudo-label selection, we incorporate uncertainty estimation (Sec.~\ref{uncertanity_and_loss_scale}) using the Monte Carlo (MC) dropout method~\cite{gal2016dropout}. This enhances the impact of pseudo-labels with low uncertainty while mitigating the effect of unreliable predictions.

Finally, we apply the feature modulator \( \mathcal{M} \) during inference to minimize domain-specific information in test data, ensuring better generalization across unseen domains.  
   
\par
\subsubsection{Similar Average Representations (SAR)}\label{Similar Average Representations}
To train the classifier to distinguish between closely located classes regardless of domain differences, we introduce SAR as a representative of such classes in the feature space. Recent works~\cite{Galappaththige_2024_CVPR,wang2023better} leverage domain-aware class prototypes to mitigate misclassification during pseudo-labeling, which often arises due to domain shifts. In contrast, our approach extends beyond pseudo-labeling by utilizing class prototypes to suppress domain-specific information during both training and testing.  

We model each source image as a combination of class-specific information—essential for classification—and domain-specific information, such as background, colors, and texture. For instance, images from the photo domain contain rich color and texture details, whereas the sketch domain provides a more abstract representation with minimal color and texture. By averaging features across multiple domains, we reduce the influence of domain-specific attributes. Recent work \cite{Galappaththige_2025_WACV} shows that averaging feature vectors leads to a more compact representative class embedding compared to alternative methods.  

Unlike existing methods that compute domain-aware class prototypes based on domain labels, we define class prototypes \( \mathbf{P} \) solely based on class labels, making them domain-agnostic. Given that \( d^L = \{(x^d, y^d)\} \) denotes the set of labeled data, we denote features \( \mathrm{K}_c \) of all labeled samples belonging to class  \( c \) across all domains as follows:
\begin{equation}
\mathrm{K}_c = \{f(x_i) \mid y_i = c\}_{i=1}^{|d^L|}
\end{equation}
where \( f \) is the feature extractor. The class prototype \( \mathbf{P}_c \) , of each class \( c \in \{1,2, \dots, C\} \), is  obtained by averaging \( \mathrm{K}_c \):
\begin{equation}\label{eq:P_c}
\mathbf{P}_c = \frac{1}{|\mathrm{K}_c|} \sum_{i=1}^{|\mathrm{K}_c|} \mathrm{K}_c[i].
\end{equation}

Next, we compute the cosine similarity between class prototypes to obtain a similarity matrix \( \mathbf{Sim} \). Using this matrix, we derive the SAR for each class \( c \), denoted as \( \mathbf{R}_c \). This is achieved by weighing each prototype \( \mathbf{P}_j \) according to its similarity \( \mathrm{Sim_{cj}}\) (see Fig.\ref{fig:method}.a) with class \( c \), ensuring that SAR is more influenced by most similar classes, where  \( \mathbf{P}_j \) denotes the class prototype of class \( j \) :
\begin{equation}\label{eq:SAR}
\mathbf{R}_c = \frac {\sum_{j=1}^{C} \mathrm{Sim_{cj}} \cdot \mathbf{P}_j} {\sum_{j=1}^{C} \mathrm{Sim_{cj}}}.
\end{equation}

We use SAR in the feature modulation process (see Fig.~\ref{fig:method}), where instance-specific features \( z \) are adjusted towards their respective similar class prototypes (Fig.~\ref{fig:intrduction_image}). During training, as the classifier improves in separating classes, inter-class similarity naturally decreases. Consequently, the contribution of other class prototypes to SAR diminishes, leading to a gradual convergence of \(\mathbf{R}_c\) towards \(\mathbf{P}_c\).

\subsubsection{Feature Modulation}
\label{Feature Modulation}

The motivation behind this modulation is similar to how the human brain reacts to an unseen target: unfamiliar features are interpreted by replacing them with familiar ones. In the same way, feature modulation modifies the original domain-sensitive features by mixing them with a portion of SARs. Since SARs are a modification of  \( \mathbf{P_c} \), which is created by averaging features from multiple source domains, they contain less domain-specific information. We introduce a feature-modulating matrix \( \mathbf{M} \), which is a learnable matrix that trains to highlight class-relevant features while down-weighting domain information containing features. 

During training, we shift the features toward SARs and force the classifier to predict the true class. This way, the feature extractor achieves better class separation while enabling the classifier to distinguish between classes more effectively, even when their embeddings are close. This process also suppresses domain information, preventing the classifier from being misled by domain variations. \( \mathbf{M} \) learns with the use of our novel diagonal loss (Eq \ref{eq:Ld},\ref{eq:Lud}). Feature modulation is applied to all source domain data, reducing the transfer of domain information to classifier (Fig.~\ref{fig:method}).

To minimize domain influence, it is important to detect features that vary greatly across domains, as these are more domain-specific. We initialize the feature-modulating matrix \( \mathbf{M} \) using the variance of the feature set \( \mathrm{K}_c \) for each class \( c \) in the labeled dataset. Initially, Features with higher variance are given lower initial weights, ensuring they contribute less to the final representation, but this \( \mathbf{M} \) learns during training, as we can't exactly define the domain information only using the labeled data. Let 
\[
\mathbf{V} = [\mathbf{V_1}, \mathbf{V_2}, \dots , \mathbf{V_C}],
\]
where \( \mathbf{V}_c = \text{Variance}(\mathrm{K}_c) \) for class \( c \). We normalize \( \mathbf{V} \) to the range \([0,1]\) and initialize \( \mathbf{M} \) as:
\begin{equation}\label{eq:M initialize}
\mathbf{M} = \mathbf{1} - \frac{\mathbf{V}- \min(\mathbf{V})}{\max(\mathbf{V})-\min(\mathbf{V})}.
\end{equation}
Tab.~\ref{tab:feature modulator} shows that this initialization outperforms random (normal distribution) initialization. During training, \( \mathbf{M} \) learns to further suppress features rich in domain information. This improves the reliability of pseudo-labels, enhancing the use of unlabeled data. 

The modulated feature representation \( \mathbf{Z}_m \) is defined as:
\begin{equation}\label{eq:modulator}
\mathbf{Z}_m = \mathcal{M}(\mathbf{Z}) = \mathbf{M} \odot \mathbf{Z} + (\mathbf{1}-\mathbf{M})\odot \mathbf{R},
\end{equation}
where \( \mathbf{Z} \) replicates the instance-specific feature \( z \) for each class \( C \), and \( \mathbf{R} \) is the SAR from Sec.~\ref{Similar Average Representations}. Here, \( \mathbf{M} \odot \mathbf{Z} \) reduces domain information, while \( (\mathbf{1}-\mathbf{M}) \odot \mathbf{R} \) shifts features closer to SARs. Each row of \( \mathbf{M} \) controls how much instance-specific information contributes to \( \mathbf{Z}_m \) for each class (Fig.~\ref{fig:method}). If a feature carries strong domain bias, its weight in \( \mathbf{M} \) is close to zero; if it is class-specific, its weight is higher, preserving its role in the modulated features.  
In practice, we identify feature positions with high domain sensitivity through \( \mathbf{M} \). Features with small values in \( \mathbf{M} \) are adjusted by adding a portion, (\( \mathbf{1-M} \)) of SARs, which strengthens class-related signals while reducing domain information. This effect is most pronounced when modulating features toward their correct class, leading to better class separation and more robust classification.

\subsubsection{Pseudo-Labeling using Modulated Features} \label{Pseudo labeling}
\par To generate more accurate pseudo-labels for unlabeled samples, we modulate the features of unlabeled data points to each class using Eq.~\ref{eq:modulator} (see Fig.~\ref{fig:method}). Following standard SSDG practice~\cite{zhou2023semi,Galappaththige_2024_CVPR,Galappaththige_2025_WACV}, we construct the unlabeled dataset \( U = d^L \cup d^U \), where weakly augmented unlabeled samples are considered alongside labeled samples without using their labels.

For a given unlabeled sample \( u_i \), we compute a modulated feature representation for each class \( c \) as \( \mathcal{M}(f(u_i), \mathbf{R}) \). Since shifting the features towards the SAR of class \( c \) mitigates domain-related information for that class, the mitigation of domain information is more effective when applied to an unlabeled data of class \(c\). So, the classifier achieves the highest prediction confidence when the features are modulated toward the true class. To identify this class, we pass the modulated feature for each class \( c \) through the classifier and compute the softmax scores:
\begin{equation}
S_i = \mathrm{softmax} ( \mathcal{C} (  \mathcal{M}(f(\alpha(u_i)),\mathbf{R}))),
\end{equation}
where \( \alpha \) denotes weak augmentation. The \( j^{th} \) row of \( S_i \) (see Fig.\ref{fig:method}.b) represents the prediction probabilities for each class after modulating the features of \( u_i \) toward class \( j \).  

To determine the predicted class, we extract the diagonal values of \( S_i \), denoted as \( \mathrm{diag}(S_i) \), which forms an array of size \( C \) containing the confidence scores for each class. The highest confidence score $p_{\mathrm{max}} = \mathrm{max} (\mathrm{diag}(S_i))$

We assign the class corresponding to \( p_{\mathrm{max}} \) as the pseudo-label \( \mathit{\Tilde{y}_i} \) for \( u_i \):
$\mathit{\Tilde{y}_i} = \mathrm{arg max}(\mathrm{diag}(S_i)).$
This approach leverages the classifier’s predictions on modulated features, which contain reduced domain-specific information, leading to more reliable pseudo-labels. As illustrated in Fig.~\ref{fig:bar_graph}, this process improves the accuracy of pseudo-labeling, enhancing the effectiveness of the SSL setting.

\subsubsection{Uncertainity Estimation of PLs and Loss Scaling}\label{uncertanity_and_loss_scale}

Unlike the fixed pseudo-label threshold used in FixMatch~\cite{sohn2020fixmatch}, we estimate the uncertainty score (US) of pseudo-labels to introduce an adaptive threshold that considers both prediction probability and uncertainty. To achieve this, we integrate dropout layers into the feature extractor, applying a dropout probability of 0.05 after the last batch normalization layer, following~\cite{kim2023use}. 

To quantify uncertainty, we perform Monte Carlo (MC) sampling by passing each input through the network multiple times with dropout enabled. We obtain a distribution of 5 predictions for each sample and compute the mean \( p_{\mathrm{max}} \) and standard deviation \( \sigma \) of the prediction probabilities.

Increasing the utilization of unlabeled data enhances domain generalization by mitigating model overfitting. However, existing methods discard a significant portion of unlabeled data during training, as they only consider predictions that exceed a high confidence threshold (typically 0.95) for pseudo-labeling. Additionally, the contribution of these pseudo-labels to training remains uniform, regardless of their prediction probability.

To address this limitation, we introduce a loss scaling function \( \mathrm{Q}(x) = \mathrm{exp}(x^3 - 1) \) that dynamically adjusts the contribution of each pseudo-label based on its prediction probability \( p_{\mathrm{max}} \) (Sec.~\ref{Pseudo labeling}). This function ensures that higher-confidence pseudo-labels have a greater influence on training while still allowing lower-confidence pseudo-labels to contribute in a controlled manner.

Our feature modulation approach improves the accuracy of pseudo-labels, enabling us to lower the confidence threshold (0.75) and effectively leverage pseudo-labels with lower confidence scores. By incorporating uncertainty estimation, we refine the pseudo-labeling process, ensuring that samples with high uncertainty are assigned labels more conservatively. This adaptive mechanism allows us to extract useful information from previously unutilized data, further enhancing the semi-supervised learning process.
\begin{equation}\label{eq:lscale}
l_{\mathrm{scale}} = \mathbbm{1}\{p_{\mathrm{max}} - \sigma > \tau\} \cdot \mathrm{Q}(p_{\mathrm{max}}).
\end{equation}

Here, \( l_{\mathrm{scale}} \) determines whether a pseudo-label is used in training based on its confidence-adjusted probability, incorporating both the mean prediction probability \( p_{\mathrm{max}} \) and uncertainty measure \( \sigma \). This approach ensures that pseudo-labels with lower confidence but high uncertainty value still contributes to the training, leading to more effective utilization of unlabeled data.

\begin{table*}[] 
\centering
\scriptsize
\begin{tabular}{@{}lcccc|cccr@{}}
    \toprule
    \multirow{2}{*}{\textbf{Method}} & \multicolumn{4}{c}{\textbf{5 labels per class} } & \multicolumn{4}{c}{\textbf{10 labels per class}} \\
    \cmidrule(lr){2-5} \cmidrule(lr){6-9}
    & \textbf{PACS} & \textbf{OfficeHome} & \textbf{VLCS} & \textbf{DigitsDG} & \textbf{PACS} & \textbf{OfficeHome} & \textbf{VLCS} & \textbf{DigitsDG}\\
    \midrule
    FixMatch & $73.4_{\pm 1.3}$ & $55.1_{\pm 0.5}$ & $69.9_{\pm 0.6}$ & $56.0_{\pm 2.2}$ & $76.6_{\pm 1.2}$ & $57.8_{\pm 0.3}$ & $70.0_{\pm 2.1}$ & $66.4_{\pm 1.4}$ \\
    FlexMatch & ${69.5}_{\pm1.1}$ & ${50.8}_{\pm0.4}$ & $56.0_{\pm 1.1}$ & $56.5_{\pm 1.8}$ & ${72.7}_{\pm1.2}$ & ${53.7}_{\pm 0.7}$ &${56.2}_{\pm 2.1}$ & $68.9_{\pm 1.2}$ \\
    StyleMatch & $\mathbf{78.4}_{\pm1.1}$ & $\underline{56.3}_{\pm0.3}$ & $72.5_{\pm 1.3}$ & $55.7_{\pm 1.6}$ & $\mathbf{79.4}_{\pm0.9}$ & $\underline{59.7}_{\pm 0.2}$ &${73.3}_{\pm 0.6}$ & $64.8_{\pm 1.9}$ \\
    FBCSA & $77.3_{\pm 1.1}$ & $55.8_{\pm 0.2}$ & $71.3_{\pm 0.7}$ & $\mathbf{62.0}_{\pm 1.5}$ & $78.2_{\pm 1.2}$ & $59.0_{\pm 0.4}$ & $72.2_{\pm 1.0}$ & $\underline{70.4}_{\pm 1.4}$  \\
    DGWM &$\underline{77.9}_{\pm0.8}$ & $56.2_{\pm 0.2}$ & $\underline{75.2}_{\pm0.9}$ & $57.4_{\pm 1.5}$ & $78.4_{\pm 1.0}$ & $59.7_{\pm 0.3}$ & $\underline{75.2}_{\pm0.7}$ & $68.4_{\pm 1.5}$\\
    \midrule
    FM (Ours) &$75.4_{\pm 1.6}$ & $\mathbf{57.2}_{\pm0.3}$ & $\mathbf{75.3}_{\pm 0.8}$ & $\underline{59.7}_{\pm1.9}$ & $\underline{78.7}_{\pm 1.0}$ & $\mathbf{61.0}_{\pm0.3}$ &$\mathbf{75.5}_{\pm0.1}$ & $\mathbf{73.0}_{\pm1.3}$\\
    \bottomrule
\end{tabular}%
\caption{Averaged top-1 accuracies over 5 independent trials in 5 labels and 10 labels setting for the major DG datasets. We achieve 3.3\% and 4.3\% increase in averaged accuracy between datasets 
for 10 labels and 5 labels settings over the FixMatch baseline respectively. (Average over 5 independent seeds is reported.) Note that StyleMatch \cite{zhou2023semi}, FBCSA \cite{Galappaththige_2024_CVPR}, and DGWM \cite{Galappaththige_2025_WACV} utilize domain labels of unlabeled data and they are unable to perform well when there are no domain labels.}
\vspace{-0.4cm}
\label{tab:all datasets}
\end{table*}

\subsection{Training}

The training process consists of four loss terms, supervised loss $L_s$, unsupervised loss $L_u$, diagonal maximizing supervised loss $L_{d}$, and diagonal maximizing unsupervised loss $L_{ud}$.
Supervised loss trains the model to give a correct prediction for weakly augmented labeled data $x_i$, $\alpha (x_i)$. We get prediction  as  ${y_{\mathrm{pred}}} = \mathrm{argmax} (\mathrm{diag}(S_i))$ where  $S_i = \mathrm{log\_softmax} ( \mathcal{C} (  \mathcal{M}(f(\alpha(x_i),\mathbf{R})))$. With $y_i$ denoting the correct label for $x_i$, supervised loss 
         $L_s = \mathrm{NLL}({y_{\mathrm{pred}}}, y_i).$
\par
 Unsupervised loss trains the model to predict the same label as the pseudo-label $\Tilde{y_i}$ for strongly augmented unlabeled data $u_i$, $\mathcal A(u_i)$. We get prediction as  ${y_{\mathrm{upred}}} = \mathrm{argmax} (\mathrm{diag}(S_{ui}))$, where  $S_{ui} = \mathrm{log\_softmax} ( \mathcal{C} (  \mathcal{M}(f(\mathcal A(u_i),\mathbf{R})))$. Then with $l_{scale}$ (Eq.\eqref{eq:lscale}) unsupervised loss, $L_u = \mathrm{NLL}({y_{\mathrm{upred}}}, \Tilde{y}_i) * l_{\mathrm{scale}}$

According to our concept of feature modulation, we emphasize the class-specific features while de-emphasizing the domain information related to class $c$. So, the classifier should give the maximum prediction for a respective class when the features are modulated to that class. We introduce diagonal maximizing loss to minimize the difference between the maximum prediction for each class $\mathrm{col_{max}}$ and the prediction for that class from the modulated features to their true class $\mathrm{diag}(S_i)$. This loss mainly trains the feature-modulating matrix $M$ to emphasize class-specific features and de-emphasize domain information containing features.

For labeled data, we define $L_{d}$ using $S_i$ computed with labeled data $x_i$ and $\mathrm{col_{\mathrm{max}}}$, where $\mathrm{col_{\mathrm{max}}}= \text{Maximum of each coloum in } S_i$ (see Fig.~\ref{fig:method}).
\begin{equation}\label{eq:Ld}
L_{d} = \mathrm{MSE}(\mathrm{diag}(S_i),\mathrm{col_{\mathrm{max}}}).
\end{equation}
\par 
For unlabeled data, we define $L_{ud}$ using $S_{ui}$ computed with unlabeled data $u_i$ and $\mathrm{col_{\mathrm{umax}}}$, where $\mathrm{col_{\mathrm{umax}}}= \text{Maximum of each coloum in } S_{ui}$ (see Fig.~\ref{fig:method}).
\begin{equation}\label{eq:Lud}
L_{ud} = \mathrm{MSE}(\mathrm{diag}(S_{ui}),\mathrm{col_{\mathrm{umax}}}) * l_{\mathrm{scale}}.
\end{equation}
This loss effectively trains the modulating matrix $\mathbf{M}$ to optimally weigh the contribution of instance-specific features and the SAR to the final modulated feature.

The final loss $L= L_{s} + L_{u}+ \beta L_d  + \gamma L_{ud},$ where $\beta$ and $\gamma$ denotes hyperparameters. We attained consistent results for $\beta$ = 1 $\gamma$ = 0.5 . See suppl. for more details.

\subsection{Inference}
During the inference, we perform feature modulation to mitigate the effect of unseen domain information. We feed the modulated features of test images $x^t$. When modulating to class c, we mitigate effect of domain information related to class c as the modulating matrix is already trained to blend more weight of domain related information from SARs. So, the domain information of unseen domain get blended by a known generalized features, reducing the effect of domain shift. Finally, we take the maximum prediction out of them as the final prediction 
$ \mathit{y_{\mathrm{pred}}^{t}} = \mathrm{argmax} (\mathit{D}^{t}),
$\text{where}$ \quad \mathit{D}^{t} = \mathrm{diag}(\mathrm{softmax} ( \mathcal{C} ( \mathcal{M} ( \mathit{f}(x^{t}))))).$


\section{Experiments}
\label{sec:Experiments}

\textbf{Datasets:} We conducted experiments on four widely used DG datasets: PACS \cite{li2017deeper}, OfficeHome \cite{venkateswara2017deep}, DigitsDG \cite{zhou2020deep}, and VLCS \cite{torralba2011unbiased}. The PACS dataset emphasizes model robustness against significant style shifts, particularly from realistic photos to abstract sketches and cartoons. OfficeHome \cite{venkateswara2017deep} contains more complexity in terms of object variety and background, offering a more challenging evaluation of generalization capabilities. In the VLCS \cite{torralba2011unbiased} dataset, the class imbalance is prominent and the performance of the proposed method proves its robustness for class imbalance. DigitsDG \cite{zhou2020deep} is a synthetic DG dataset with digit classification tasks across four domains. See suppl. for a detailed description of datasets and details of the class imbalance.

\noindent \textbf{Training and Implementation Settings:}
We conducted our training following the same training settings as in StyleMatch \cite{zhou2023semi}, FBCSA \cite{Galappaththige_2024_CVPR} and DGWM \cite{Galappaththige_2025_WACV}. We used ImageNet \cite{deng2009imagenet} pretrained ResNet-18 \cite{he2016deep} as the feature extractor for all the experiments. We followed leave-one-domain-out training protocol which is widely used in DG and SSDG methods. Our experiments were conducted in 5-label and 10-label settings. A learning rate of 0.03 was applied across all runs, with training completed over 20 epochs for each dataset. These learning rates are decayed using cosine annealing. For mini-batch construction, we randomly sampled 16 labeled and 16 unlabeled samples per source domain. We report top-1 accuracy averaged over 5 independent trials.

 
\noindent \textbf{Main Results}
As shown in Tab \ref{tab:all datasets}, our method achieves consistent improvements of \textbf{3.3}\% and \textbf{4.3\%} in averaged accuracy across datasets compared to the FixMatch \cite{sohn2020fixmatch} for 10 and 5 labels settings, respectively. Our method has notable gains in OfficeHome, VLCS and DigitsDG, surpassing all the SOTA methods. 
\textbf{PACS:} We obtain 2\% and 2.1\% average gains over FixMatch for 5 and 10 labels settings, respectively. \textbf{OfficeHome:} We attain the best accuracies of 57.2\% and 61.0\% for 5 and 10 labels setting. Our approach significantly surpasses all the SOTA methods for both settings. \textbf{VLCS:} Overcoming the challenge of class imbalance in the VLCS dataset, we attain a gain of 5.4\% compared to FixMatch in both settings. Furthermore our method outperforms all the SOTA methods with accuracies of 75.5\% and 75.3\% for 5 and 10 labels settings respectively. \textbf{DigitsDG:} Our approach exceeds FixMatch by a significant gain of 6.6\% and 3.7\% for 10 and 5 labels settings respectively.

\begin{table}[]
\scriptsize
\centering
\begin{tabular}{@{}p{3.5cm} c c c r@{}}
\hline
\textbf{Method} & $\tau$ & \textbf{Acc.} & \textbf{Keep Rate} & \textbf{PL Acc.} \\
\hline

Baseline FixMatch & 0.95 & 57.8 &  63.5 & 87.1 \\
Baseline + Loss Scaling & 0.95 & 60.2 & 52.2 & \underline{88.3}\\
Baseline + Feature Modulator ($\mathcal{M}$) &  0.75  & \underline{60.8} & \underline{78.5} & 85.0 \\
Baseline + Uncertainty Score (US)  &   0.75 & 58.5 & 63.7 & 87.6 \\
Baseline + Loss Scaling + $\mathcal{M}$  & 0.75   & 60.6  & 75.8 & 87.7 \\
Baseline + Loss Scaling + US   &   0.75 & 57.7 & 76.4 & 84.7 \\
Baseline + $\mathcal{M}$ + US   &  0.75  & 57.4 & \textbf{79.3} & 82.4 \\
Baseline + $\mathcal{M}$ + Loss Scaling + US   &  0.75  & \textbf{61.0} & 67.6 &  \textbf{90.4}\\

\hline
\end{tabular}
\vspace{-0.1cm}
\caption{Contribution of key components; Loss scaling alone degrades performance due to noisy pseudo-labels, especially with lower confidence thresholds. The feature modulator improves accuracy by accurate pseudo-labels but has a low keep rate, indicating poor unlabeled data use. Combining both enhances pseudo-label accuracy and unlabeled data utilization.}
\vspace{-0.1cm}
\label{tab:key components}
\end{table}
\begin{table}[]
\scriptsize
\centering
\begin{tabular}{@{}p{3.5cm} c c r@{}}
\hline
\textbf{Feature Modulator ($\mathcal{M}$)} & \textbf{Accuracy} & \textbf{Keep Rate} & \textbf{PL Accuracy} \\
\hline

Normal Init. + SAR + $L_d$ + $L_{ud}$  & 47.8 & 62.3 & 87.2 \\
Variance Init. + AP + $L_d$ + $L_{ud}$  & \underline{60.5} & \textbf{80.5} & 85.3  \\
Variance Init. + SAR  &  60.4 & 66.7 & \underline{90.3} \\
Variance Init. + SAR + $L_d$ + $L_{ud}$ & \textbf{61.0} & \underline{67.6} & \textbf{90.4} \\
\hline
\end{tabular}
\vspace{-0.1cm}
\caption{Comparison of feature modulator components at $\tau$ = 0.75. Variance initialization outperforms normal initialization, SAR outperforms average prototypes (AP), and diagonal loss is imperative for feature modulator.}
\vspace{-0.1cm}
\label{tab:feature modulator}
\end{table}
\begin{table}[]
\scriptsize
\centering
\begin{tabular}{@{}p{2cm} c c c r@{}}
\hline
\textbf{Conf. Thresh. ($\tau$)} & \textbf{Loss Scaling} &  \textbf{Accuracy} & \textbf{Keep Rate} & \textbf{PL Acc.} \\
\hline
0.95& \ding{55} & 59.7 & 41.9 & \textbf{95.9}\\
0.95& \ding{52} & 59.8 & 41.6 & \textbf{95.9} \\
0.80& \ding{52} & 60.6 & 63.2 & \underline{91.6}  \\
0.75& \ding{55} & {57.4} & \textbf{79.3} & 82.4 \\
0.75& \ding{52} & \textbf{61.0} & {67.6} & 90.6 \\
0.70& \ding{52} & \underline{60.8} & \underline{71.2} & 89.4 \\
\hline
\end{tabular}
\vspace{-0.1cm}
\caption{Comparison of our method with various confidence thresholds with and without loss scaling. Experiments show that incorporating loss scaling with a 0.95 confidence threshold limits unlabeled data utilization, causing sub-optimal performance. The optimal balance occurs at a 0.75 threshold; further lowering thresholds degrades the performance.}
\vspace{-0.3 cm}
\label{tab:loss scaling with various threholds}
\end{table}

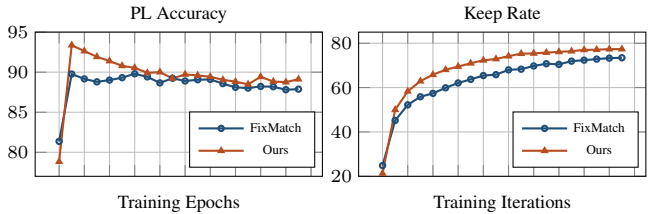
\begin{figure}[h]
     \centering

\definecolor{fixmatchLine}{HTML}{1F4E79}  
\definecolor{dgwmLine}{HTML}{4A7023}      
\definecolor{oursLine}{HTML}{B84E20}      
\scriptsize
\begin{tikzpicture}
    \begin{axis}[
        at={(4.3cm,0)}, 
        width=5.4cm,
        height=3.5cm,
        ylabel={},
        ymin=20, ymax=85,
        xtick={1,2,3},
        xtick={1,3,...,19, 21},
        xticklabels={},
        title={Keep Rate},   
        title style={yshift=-5pt}, 
        xlabel={Training Iterations},
        grid=major ,
        legend style={
            at={(0.975,0.45)}, 
            /tikz/every node/.append style={align=center, font=\tiny}, 
        }, 
    ]
        \addplot[color=fixmatchLine, mark=o, mark options={scale=0.5},thick] coordinates {(1, 24.84) (2, 45.20) (3, 52.23) (4, 55.86) (5, 57.39) (6, 59.90) (7, 62.07) (8, 63.72) (9, 65.41) (10, 65.82) (11, 67.99) (12,68.29 ) (13, 69.74) (14,70.75 ) (15,70.46 ) (16, 71.91) (17, 72.36) (18, 72.84) (19, 73.27) (20,73.44)};
        \addplot[color=oursLine, mark=triangle,mark options={scale=0.5}, thick] coordinates {(1,21.31) (2, 49.98 ) (3, 58.33 ) (4, 62.89) (5, 65.81) (6, 68.03 ) (7, 69.49 ) (8, 70.95) (9, 72.3 ) (10, 72.9) (11, 74.16 ) (12, 75.29) (13, 75.41 ) (14, 75.77 ) (15, 76.1) (16, 76.4) (17,77.0 ) (18,77.07 ) (19, 77.34) (20,77.35)};
        \legend{FixMatch, Ours}

    \end{axis}
    \begin{axis}[
        width=5.4cm,
        height=3.5cm,
        xlabel={Training Iterations},
        ylabel={},
        ymin=77, ymax=95,
        xtick={1,2,3},  
        xtick={1,3,...,19,21},
        xticklabels={}, 
        title={PL Accuracy},
        title style={yshift=-5pt}, 
        grid=major,
        legend style={
            at={(0.975,0.45)}, 
            /tikz/every node/.append style={align=center, font=\tiny}, 
        },   
    ]
        \addplot[color=fixmatchLine, mark=o, mark options={scale=0.5}, thick] coordinates {(1,81.36 ) (2, 89.77 ) (3, 89.16 ) (4, 88.79) (5, 89.01) (6, 89.31 ) (7, 89.80 ) (8, 89.41) (9, 88.66 ) (10, 89.24) (11, 88.90 ) (12, 89.05) (13, 89.10 ) (14, 88.57 ) (15, 88.11) (16, 88.01) (17,88.22 ) (18,88.18 ) (19, 87.81) (20,87.88)};
        \addplot[color=oursLine, mark=triangle,mark options={scale=0.5}, mark options={scale=0.5}, thick] coordinates {(1,78.83) (2, 93.37 ) (3, 92.63 ) (4, 91.92) (5, 91.39) (6, 90.79 ) (7, 90.54 ) (8, 89.91) (9, 90.01 ) (10, 89.22) (11, 89.72 ) (12, 89.59) (13, 89.43 ) (14, 89.04 ) (15, 88.82) (16, 88.48) (17,89.42 ) (18,88.83 ) (19, 88.76) (20,89.12)};
        \legend{FixMatch, Ours}
    \end{axis}

\end{tikzpicture}
     \vspace{-0.6cm}
      \caption{ Comparison of PL accuracy and keep rate with training iterations (20) for target domain Art of OfficeHome dataset. We achieve consistent improvements over FixMatch.}
      \vspace{-0.3cm}
     \label{fig:line_graph}
 \end{figure}
\noindent \textbf{Contribution of Key Components:}
We conducted a comprehensive analysis of the key components of the proposed method under 10 label settings for the OfficeHome dataset. (Tab \ref{tab:key components}). 
The feature modulator alone shows a considerable improvement in accuracy due to the effective emphasis of the class-related features. Even with a higher average accuracy and keep rate, it shows lower pseudo-label accuracy. This illustrates the dilemma between keep rate and pseudo-label accuracy. We combine the loss scaling and uncertainty score to the feature modulator to make the model robust with other benchmarks, balancing accurate pseudo-labels with higher unlabeled data utilization.
\newline\textbf{Feature modulator components:} We compared proposed feature variance based initialization method for feature modulating matrix $\textbf{M}$ with normal initialization in Tab \ref{tab:feature modulator}. The variance initialization has a significant impact compared to the normal initialization. 
SARs show a better improvement over the average prototypes (AP) generated by averaging all the class prototypes. In the SAR generation, the process of averaging class prototypes based on cosine similarity provides better feature modulation for clear class separation. The impact of the contribution of the diagonal losses on accurate class separation of the classifier is evident from the accuracy improvement.\\ 
\noindent \textbf{Impact of confidence threshold and loss scaling:}
Tab \ref{tab:loss scaling with various threholds} 
shows performance across thresholds with and without loss scaling, which is more effective at 0.75 than 0.95. Lowering the threshold increases inclusion but also introduces noisy pseudo-labels that degrade learning and limit Keep Rate. Thus, no sharp increase is observed. A threshold of 0.75 achieves the best balance, while further reduction harms performance.\\
\noindent\textbf{Improved PL accuracy and keep rate:} 
We plot the PL accuracy and keep rate of our method in Fig. \ref{fig:line_graph} compared to FixMatch \cite{sohn2020fixmatch}. Throughout training iterations, we achieve consistent improvements in both the PL accuracy and the keep rate. With improved average accuracy, we verify our claim of balancing the dilemma between accurate pseudo-labels and higher unlabeled data utilization.

\begin{table}[]
    \centering
    \scriptsize
    \setlength{\tabcolsep}{4pt} 
    \renewcommand{\arraystretch}{1} 
    \begin{tabular}{@{}lccc|ccc|ccr@{}}
        \toprule
        \multirow{2}{*}{\textbf{Method}} & \multicolumn{3}{c}{\textbf{D1}} & \multicolumn{3}{c}{\textbf{D1+D2}} & \multicolumn{3}{c}{\textbf{D1+D2+D3}} \\
        \cmidrule(lr){2-4} \cmidrule(lr){5-7} \cmidrule(lr){8-10}
        & \textbf{Acc.} & \textbf{KR} & \textbf{PL Acc.} & \textbf{Acc.} & \textbf{KR} & \textbf{PL Acc.} & \textbf{Acc.} & \textbf{KR} & \textbf{PL Acc.} \\
        \midrule
        FixMatch & \textbf{59.5} & 37.0 & \textbf{93.7} & 63.6 & 49.1 & \textbf{89.7} & 67.2 & 62.6 & 88.2 \\
        FM (ours) & 59.0 & \textbf{41.1} & 92.1 & \textbf{64.2} & \textbf{53.9} & 89.1 & \textbf{71.2} & \textbf{63.4} & \textbf{90.7} \\
        \bottomrule
    \end{tabular}
    \vspace{-0.1cm}
    \caption{Performance of our method with FixMatch for the increase of source domains (D1-Art, D2-ClipArt, D3-Product) of OfficeHome where target domain is real world. Our method maintains higher average accuracy, while the baseline's drops.}
    \vspace{-0.6cm}
    \label{tab:source_domain_increase}
\end{table}

\section{Conclusion}
We address a unique problem in relatively under-explored research area SSDG, where domain labels are unavailable for model training.
We address the \textit{dilemma}
between more accurate pseudo-labels and a higher keep rate. Existing models struggle to achieve better data utilization with higher pseudo-label accuracy.
The proposed feature modulation improves the generalization by pushing the features towards similar average representations and loss scaling improves unlabeled data utilization by effectively using low-confident pseudo-labels. This encourages the classifier to accurately classify closely clustered features despite the domain by minimizing the propagation of domain information. Extensive experiments with major DG benchmark datasets validate our claims.
{
 \small
\bibliographystyle{ieeenat_fullname}
    \bibliography{main}
}
\clearpage
\setcounter{page}{1}
\maketitlesupplementary


\begin{figure}[]
\scriptsize
     \centering
     \includegraphics[width=0.48\textwidth]{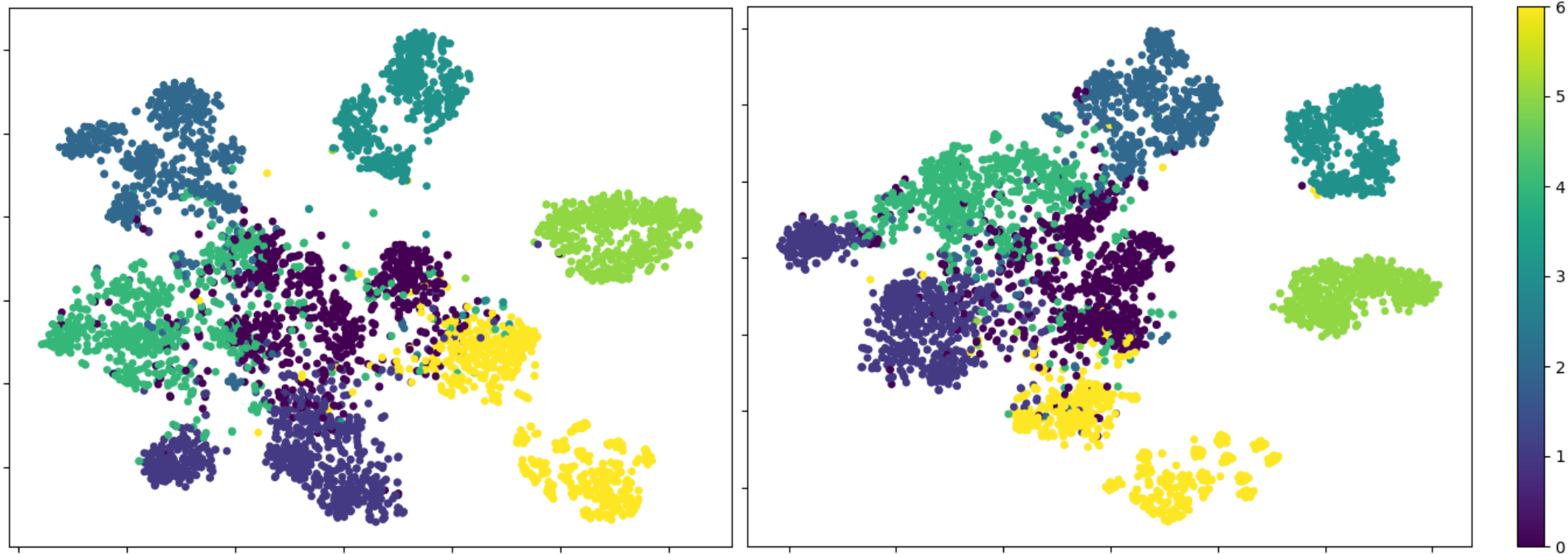}
      \caption{t-SNE visualization of features of three source domains (Cartoon, Photo, and Sketch) in the PACS dataset after training. The separation has gained improvements for class labels 0, 2, 4, and 5. (dog, giraffe, horse, and house respectively).
      {\bf Left}: Features from feature extractor in FixMatch. 
      {\bf Right}: Modulated features in our method.
      }
     \label{fig:tsne1}
 \end{figure}

\begin{figure}[]
\scriptsize
     \centering
     \includegraphics[width=0.48\textwidth]{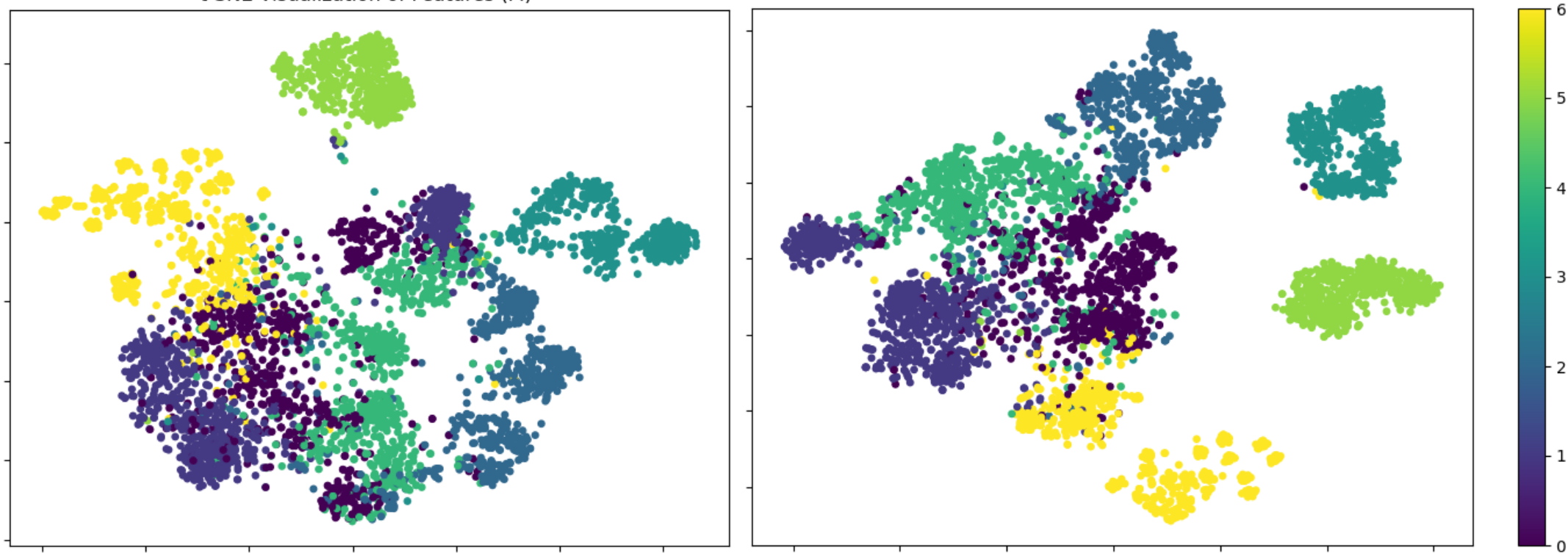}
      \caption{t-SNE visualization of features of three source domains (cartoon, photo, and sketch) in the PACS dataset at the beginning and end of the training. Overall the separation has gained improvements for all class labels. With the training iterations, the model has sufficiently removed the domain-wise separation in the above classes, achieving better domain generalization.
      {\bf Left}: Modulated features at the beginning of training after the first iteration. 
      {\bf Right}: Modulated features at the end of training after the last iteration.
      }
     \label{fig:tsneEpoch1}
 \end{figure}

\begin{table*}[]
\footnotesize
\centering
\begin{tabular}{@{}p{2.5cm} c c c c r@{}}
\hline
\textbf{} &  \textbf{FixMatch} &  \textbf{StyleMatch} & \textbf{FBCSA} & \textbf{DGWM} & \textbf{Ours} \\
\hline
Domain labels used & \ding{55} &\ding{52} & \ding{52} & \ding{52} & \ding{55} \\
Confidence threshold & 0.95& 0.95 &0.95 & 0.95 & 0.75 \\
Loss scaling used  &\ding{55} &\ding{55} & \ding{55} & \ding{55} & \ding{52} \\
Prototypes used &\ding{55} &\ding{55} & Domain-aware class prototypes &\ding{55} & Class prototypes \\
\hline
\end{tabular}
\caption{Comparison of our method with SOTA.}
\label{tab:SOTA comparison}
\end{table*}

\begin{table}[]
    \centering
    \footnotesize
    \begin{tabular}{@{}lccr@{}}
        \hline
        Domain  & \# of Samples & Ratio & Highest/Lowest \\
        \hline
        \addlinespace[2pt]
        Caltech  & 809  & 62  & 13.0  \\
        LabelMe  & 1124 & 39  & 28.9  \\
        Pascal   & 1394 & 307 & 4.6   \\
        SUN      & 1175 & 19  & 61.9  \\
        \hline
    \end{tabular}
    \caption{Sample distribution and ratio statistics across domains in VLCS.}
    \label{tab:vlcs_stats}
\end{table}

\begin{table}[]
    \centering
    \scriptsize
    \setlength{\tabcolsep}{2.5pt}
    \begin{tabular}{@{}lccc|cc r@{}}
        \toprule
        \multirow{2}{*}{\textbf{Method}} & \multicolumn{3}{c}{\textbf{OfficeHome}} & \multicolumn{3}{c}{\textbf{PACS}} \\
        \cmidrule(lr){2-4} \cmidrule(lr){5-7}
        & \textbf{Accuracy} & \textbf{Keep Rate} & \textbf{PL Acc.} & \textbf{Accuracy} & \textbf{Keep Rate} & \textbf{PL Acc.} \\
        \midrule
        PL without FM & 60.7 & 66.4 & 89.8 & 78.3 & 93.9 & \textbf{93.2} \\
        FM (Ours) & \textbf{61.0} & \textbf{67.6} & \textbf{90.4} & \textbf{78.7} & \textbf{94.3} & 93.0 \\
        \bottomrule
    \end{tabular}
    \caption{Impact of feature modulator during pseudo labeling on OfficeHome and PACS datasets.}
    \label{tab:feature_modulation}
\end{table}

\begin{table*}[]
    \centering
    \scriptsize
    \setlength{\tabcolsep}{5pt} 
    \renewcommand{\arraystretch}{1.2} 
    
    \resizebox{\textwidth}{!}{  
    \begin{tabular}{@{}lcccc|cccr@{}}
        \toprule
        \multirow{2}{*}{\textbf{Method}} & \multicolumn{4}{c}{\textbf{5 labels per class} } & \multicolumn{4}{c}{\textbf{10 labels per class}} \\
        \cmidrule(lr){2-5} \cmidrule(lr){6-9}
        & \textbf{PACS} & \textbf{OfficeHome} & \textbf{VLCS} & \textbf{DigitsDG} & \textbf{PACS} & \textbf{OfficeHome} & \textbf{VLCS} & \textbf{DigitsDG}\\
        \midrule
        ERM & 51.2±3.0 & 51.7±0.6 & 67.2±1.8 & 22.7±1.0 & 59.8±2.3 & 56.7±0.8 & 68.0±0.3 & 29.1±2.9  \\
        MixUp & 45.3±3.8 & 52.7±0.6 & 69.9±1.3 & 21.7±1.9 & 58.5±2.2 & 57.2±0.6 & 69.6±1.0 & 29.7±3.1\\
        GroupDRO & 48.2±3.6 & 53.8±0.6 & 69.8±1.2 & 23.1±1.9 & 57.3±1.2 & 57.8±0.4 & 69.4±0.9 & 31.5±2.5\\
        \midrule
        ERM + PL & 62.8±3.0 & 54.2±0.6 & 65.4±2.9 & 43.4±2.9 & 63.0±1.5 & 55.5±0.3 & 60.5±1.1 & 55.0±2.4  \\
        MixUp + PL & 60.6±2.9 & 51.9±0.4 & 60.8±2.8 & 35.4±1.3 & 62.3±1.9 & 55.1±0.2 & 64.4±1.1 & 43.5±1.0 \\
        GroupDRO + PL & 62.3±1.9 & 54.5±0.5 & 69.3±0.3 & 39.4±1.3 & 62.1±2.0 & 58.5±0.3 & 66.5±0.2 & 49.9±1.9 \\
        \bottomrule
    \end{tabular}
    } 
    \caption{Comparison with the DG methods, DG+PL \cite{sohn2020fixmatch} methods under the first setting, i.e., only a few instances (5,10) are labeled from each source domain.}
    \label{DG}
\end{table*}

\renewcommand{\thesection}{\Alph{section}}
\setcounter{section}{0}

\section{Comparison with SOTA}

 Table \ref{tab:SOTA comparison} provides a comparison of our method with SOTA methods of SSDG. The SSDG problem we address is unique as it does not require domain labels for training. We have effectively reduced the confidence threshold for pseudo-labels from 0.95 to 0.75 in order to achieve better unlabeled data utilization together with better average accuracy, as in Fig.\ref{fig:bar_graph} . Loss scaling is one of the major contributions we introduce in our proposed method to leverage more unlabeled data for model training. FBCSA \cite{Galappaththige_2024_CVPR} uses domain-aware class prototypes, a variant of class prototypes related to each domain. In our work, we use class prototypes irrespective of the domain. The dimensions of {\small $S_i, z, \mathbf{Z}, \mathbf{M}, \mathbf{Z}_m$} are {\small $C \times C$, $1\times 512$, $C\times 512$, $1\times 512$, $C\times 512$, $C \times 512$} respectively.


\section{Improvement in Class Separation}
From the t-SNE visualization in Fig.~\ref{fig:tsne1}, we can empirically show the improvement in class separation of the proposed method compared to the baseline, FixMatch \cite{sohn2020fixmatch} after all training iterations. The figure shows the features of all three source domains—cartoon, photo, and sketch—of the PACS \cite{li2017deeper} dataset when the target domain is art painting.  For each class, the separation between three domains is visible, especially for class labels 0, 2, 3, and 4 in Fig.~\ref{fig:tsne1}, Left. Our method in Fig.~\ref{fig:tsne1}, Right, has reduced the domain-wise separation for better domain generalization. The features of class 0 overlap with its nearby classes in Fig.~\ref{fig:tsne1}, Left. Our method has sufficiently reduced that overlapping between nearby classes, leading to better class separation. 

Fig.~\ref{fig:tsneEpoch1} shows the learning of the proposed method with training iterations. Initially, the domain separation within each class is high, with clearly visible three clusters for each class (Fig.~\ref{fig:tsneEpoch1} Left). These separate clusters belong to separate three domains Our method in Fig.~\ref{fig:tsneEpoch1} Right, has sufficiently reduced this domain separation within classes to achieve better domain generalization. The closely located four classes (0, 1, 4, and 6) has achieved a notable separation in our method.

\begin{algorithm}
\caption{Pseudo-code}
\begin{algorithmic}[1]

\Require Labeled data ($x_i$ , $y_i$) , Unlabeled data ($u_i$,), Weakly augmented $\alpha(x_i)$, Strongly augmented $\mathcal{A}(u_i)$, Confidence threshold $\tau$, Total epochs $E$, Variance init. $\mathbf{V}$ (Sec.\ref{Feature Modulation}), Loss scaling function: $\mathrm{Q}(x)=\mathrm{exp}(x^3-1)$, Feature extractor: $f$, Feature modulator: $\mathcal{M}$, Classifier: $\mathcal C$, Model: $F = \mathcal C ( \mathcal{M} ( f))$

\State Initialize $\mathbf{M} = \mathbf{1} - \frac{\mathbf{V}- \min(\mathbf{V})}{\max(\mathbf{V})-\min(\mathbf{V})}$, $ $ Modulating matrix 

\State \# Define Feature Modulator $\mathcal{M}$($z$, $\mathbf{R}$)
\State $\mathbf{Z} = \text{Replicate } z \text{ by number of classes } C$
\State $\mathbf{Z}_m = \mathcal{M}(\mathbf{Z}) = \mathbf{M} \odot \mathbf{Z} + (\mathbf{1}-\mathbf{M}) \odot \mathbf{R}$

\For{$\mathrm{epoch} = 1$ to $E$}

    \State $\mathbf{P} = $ Compute class prototypes using Eq.\ref{eq:P_c}
    \State \#Compute similar average representations $\mathbf{R}$
    \State $\mathrm{Sim} = $ cosine similarity matrix between each $\mathbf{P_c}$
    \State $\mathbf{R_c} = \frac{\sum_{j=1}^{C} \mathrm{Sim}_{cj} \cdot \mathbf{P_j}}{\sum_{j=1}^{C} \mathrm{Sim}_{cj}}$
    
    \State $\mathbf{R} = [\mathbf{R_1}, \mathbf{R_2}, \dots , \mathbf{R_C}]$

    \State \#Generating Pseudo-Labels 
    \State $S_i = [\hspace{5pt}]$
    \For{$\mathrm{k} = 1$ to 5}
        \State $ S_{i\mathrm{k}} = \mathrm{softmax}( \mathcal{C}  (\mathcal{M}(f(u_i),\mathbf{R})))$
        \State $S_i = S_i \cup  S_{i\mathrm{k}}$
        
    \EndFor
    \State $D_{\mu}, D_{\sigma} = \mathrm{mean}(\mathrm{diag}(S_i)),  \mathrm{std}(\mathrm{diag}(S_i))$
    \State $p_{\mathrm{max}}, \mathit{\Tilde{y}_i} = \mathrm{max} (D_{\mu}), \mathrm{arg max} (D_{\mu})$
    \State $\sigma$ = std of $p_{\mathrm{max}}$
    \State $ l_{\mathrm{scale}} = [ \mathbbm{1}\{p_{\mathrm{max}} - \sigma > \tau\}$ $*$ $\mathit{Q}(p_{\mathrm{max}})$ ]

    \State \#Compute Supervised Loss
    \State $ S_i = \mathrm{log\_softmax} ( \mathcal{C} (  \mathcal{M}(f(\alpha(x_i)),\mathbf{R})))$
    \State ${y_{\mathrm{pred}}} = \mathrm{argmax} (\mathrm{diag}(S_i))$
    \State $\mathrm{col_{\mathrm{max}}}$ = Maximum of each class in $S_i$
    \State $L_s = \mathrm{NLL}({y_{\mathrm{pred}}}, y_i)$
    \State $L_{d} = \mathrm{MSE}(\mathrm{diag}(S_i),\mathrm{col_{\mathrm{max}}})$

    \State \#Compute Unsupervised Loss
    \State $ S_{ui} = \mathrm{log\_softmax} ( \mathcal{C} (  \mathcal{M}(f(\mathcal{A}(u_i)),\mathbf{R})))$
    \State ${y_{\mathrm{upred}}} = \mathrm{argmax} (\mathrm{diag}(S_{ui}))$
    \State $\mathrm{col_{\mathrm{umax}}} = $ Maximum of each class in $S_{ui}$
    \State $L_u = \mathrm{NLL} ({y_{\mathrm{upred}}}, \mathit{\Tilde{y}_i})$ $*$ $l_{\mathrm{scale}}$
    \State $L_{ui} = \mathrm{MSE}(\mathrm{diag}(S_{ui}),\mathrm{col_{\mathrm{umax}}})$ $*$ $l_{\mathrm{scale}}$ 

    \State \Return $L = L_s + L_u + \beta L_{d} + \gamma L_{ud}$

\EndFor 

\end{algorithmic}
\label{alg:algo}
\end{algorithm}

\begin{table}[h!]
\centering
\begin{tabular}{lccc}
\toprule
 & \textbf{Expression} & \textbf{OfficeHome ($c=65$)} & \textbf{PACS ($c=7$)} \\
\midrule
\(\mathcal{M}\) & $c \times 512$ & 33,280 & 3,584 \\
$F$ & 11,176,512 & 11,176,512 & 11,176,512 \\
\(\mathcal{C}\) & $512 \times c + c$ & 33,345 & 3,591 \\
\bottomrule
\end{tabular}
\caption{Number of learnable parameters for different components of proposed method in OfficeHome and PACS datasets. 
Here, $\mathcal{M}$, $\mathit{f}$, and $\mathcal{C}$ are feature modulator, feature extractor and classifier respectively}
\label{tab:params}
\end{table}

\begin{table}[h!]
\centering
\begin{tabular}{@{}lcc@{}}
    \toprule
    \textbf{Method} & \textbf{Average time/epoch} & \textbf{Overhead} \\
    \midrule
    FixMatch  & 22.5  & - \\
    FBCSA       & 36.5  & 58.22\% \\
    StyleMatch  & 68.25 & 203.33\% \\
    DGWM         & 25.5  & 13.33\% \\
    FM (Ours)         & 24.75  & 10.00\% \\
    \bottomrule
\end{tabular}
\caption{Training overhead comparison over the FixMatch baseline.}
\label{tab:training_overhead}
\end{table}

\begin{table*}[h!] 
\centering
\scriptsize
\begin{tabular}{@{}lccc|ccc|ccc|ccr@{}}
    \toprule
    \multirow{2}{*}{\textbf{LR}} 
    & \multicolumn{3}{c}{\textbf{PACS}} & \multicolumn{3}{c}{\textbf{OfficeHome}} & \multicolumn{3}{c}{\textbf{VLCS}} & \multicolumn{3}{c}{\textbf{DigitsDG}} \\
    \cmidrule(lr){2-4} \cmidrule(lr){5-7} \cmidrule(lr){8-10} \cmidrule(lr){11-13}
    & $\beta=0.5$ & $\beta=0.1$ & $\beta=1.0$ & $\beta=0.5$ & $\beta=0.1$ & $\beta=1.0$ & $\beta=0.5$ & $\beta=0.1$ & $\beta=1.0$ & $\beta=0.5$ & $\beta=0.1$ & $\beta=1.0$ \\
    & $\gamma=0.5$ & $\gamma=0.5$ & $\gamma=0.5$ & $\gamma=0.5$ & $\gamma=0.5$ & $\gamma=0.5$ & $\gamma=0.5$ & $\gamma=0.5$ & $\gamma=0.5$ & $\gamma=0.5$ & $\gamma=0.5$ & $\gamma=0.5$ \\
    \midrule
    0.02 & 77.5 & 78.0 & 78.4 & 61.0 & 61.0 & 60.7 & 75.2 & 75.3 & \textbf{75.5} & 69.1 & 71.9 & 70.1 \\
    0.03 & 77.9 & 78.1 & \textbf{78.7} & 60.3 & 60.2 & \textbf{61.0} & 75.1 & 75.2 & 75.5 & 68.7 & 71.4 & \textbf{73.0} \\
    0.04 & 78.4 & 78.3  & 78.6 & 60.7 & 60.2 & 60.5 & 75.2 & 75.5 & 75.3 & 68.5 & 71.7 & 72.0  \\
    \bottomrule
\end{tabular}
\caption{Hyperparameter tuning of learning rates and loss scales across datasets PACS, OfficeHome, VLCS, and DigitsDG.}
\label{tab:Loss scale version}
\end{table*}
\section{Computational complexity}
We compared the training efficiency of our method with the existing SSDG methods \cite{zhou2023semi,sohn2020fixmatch}. In StyleMatch \cite{zhou2023semi}, they transfer the styles of one domain to other domain images during the training and consider style-modified images as an augmentation method. This style augmentation is implemented using AdaIN. StyleMatch adds a 200\% 
additional overhead to FixMatch due to these style augmentations but our method only adds a small overhead of 10\%. We report the calculations using the average time computed for five seeds.

As shown in Tab.~\ref{tab:params}, the distribution of learnable parameters across different modules indicates that the feature extractor (\( \mathcal{F} \)) accounts for the majority of parameters, while the feature modulator (\( \mathcal{M} \)) introduces only a small overhead relative to the total size of the model. The classifier (\( \mathcal{C} \)) scales with the number of classes in each dataset, but its parameter count remains relatively modest compared to the feature extractor. This analysis highlights that the proposed feature modulation adds minimal computational cost while providing the intended benefits in suppressing domain-specific information.

\section{Description of Fig.\ref{fig:intrduction_image} - Feature Modulation \& SAR generation}
The Fig.\ref{fig:intrduction_image} \textbf{A} shows the feature space with class prototypes for each class separated by boundary lines. \textbf{B} shows how SARs are generated. Each class prototype contributes to the SAR depending on how far it is. The glow associated with each class prototype shows the cosine similarity of that class to the class $c$. The closest class prototypes have the highest glow intensity. \textbf{C} is the feature space with all the SARs denoted by green squares. The extracted features are shown as the instance-specific features, $z$. In \textbf{D}, we modulate $z$ toward every SAR by a weight that is learnt during training. We represent these modulated features in gray color stars.    

\newcommand{\insertpython}[2]{
    \lstinputlisting[style=python, caption={#2}]{#1}
}

\section{Why are domain labels a luxury?}
The field of semi-supervised learning has emerged due to the practical challenge of obtaining completely labeled data, as acquiring class labels is often a luxury. Similarly, the presence of domain labels is a luxury because the process of getting domain labels is equally difficult. A t-SNE or a clustering of domain generalization datasets does not separate domains clearly. As shown in \ref{fig:tsne1}, t-SNE does not achieve clear domain separation for some classes, and domain mixing further complicates separation. Hence, we consider the presence of domain labels as a luxury, which most of the SOTA SSDG methods rely on \cite{Galappaththige_2024_CVPR, Galappaththige_2025_WACV}. We propose an SSDG method which does not require domain labels.

\section{Performance when SARs are generated only from classes with high similarity}
We make the model more generalized by shifting the feature spaces towards SAR, which explained in the Sec.\ref{Similar Average Representations}. The goal is to show the worst case scenarios with closely related classes that the classifier can encounter during the testing. We compute the SAR for each class by considering the similarity of that class with other classes. 

 As an alternative method, we experiment our method by creating the SAR in a different way. Here we get the average of the prototypes most similar classes to a respective class. In this way we decided on a similarity threshold and get the prototypes that have a higher similarity than that threshold (set to 0.87 after testing). 
We validate that the SAR generation is more effective when SARs are generated using a similarity matrix using the results shown in table \ref{tab:feature modulator}
\section{Detailed description of datasets}
We conducted experiments on four widely used DG datasets: PACS \cite{li2017deeper}, OfficeHome \cite{venkateswara2017deep}, VLCS \cite{torralba2011unbiased}, and DigitsDG \cite{zhou2020deep}. The OfficeHome \cite{venkateswara2017deep} dataset consists of images from four distinct domains: Art, Clipart, Product, and Real-World, covering a total of 65 classes with 15,500 images. The PACS \cite{li2017deeper} dataset features images from four domains---art painting, cartoon, photo and sketch---, and it includes 9,991 images and 7 common classes in total. The VLCS \cite{torralba2011unbiased} dataset aggregates images from four major public datasets—Pascal VOC 2007 (V), LabelMe (L), Caltech-101 (C), and Sun09 (S), with five shared classes. VLCS is useful for domain generalization as each dataset in VLCS serves as a distinct domain with different data collection sources and environments. In addition, the VLCS dataset can be used to evaluate the robustness to class imbalance. DigitsDG \cite{zhou2020deep} is a synthetic domain generalization dataset with digit classification tasks across four domains: MNIST, MNIST-M, SVHN, and SYN. 
All of these benchmark datasets make domain generalization a more challenging task generalization due to the significant visual differences between domains such as varying color schemes and background textures.

\section{Performance under class imbalance}

The class imbalance is prominent in the VLCS dataset as shown in Table \ref{tab:vlcs_stats}. The experiments in Table \ref{tab:all datasets} with the VLCS dataset prove that our method performs well under class imbalance. Our method has a significant accuracy improvement of 5.4\% and 5.4\% for the setting of 5 labels and 10 labels, respectively in VLCS dataset.

\section{Impact of feature modulation during pseudo labeling}
The purpose of the feature modulation is to reduce the effect of domain-related features on the classification process. In our method we used the feature modulation during the pseudo label process to improve the accuracy of pseudo labels by mitigating the domain effect. We validate our claim by the following experiments shown in Table \ref{tab:feature_modulation} which includes the pseudo label accuracies and keep rates in the presence and absence of feature modulation during the pseudo labeling process.

Table \ref{tab:feature_modulation} presents an ablation study evaluating the effect of the feature modulator on pseudo-label generation. Feature modulation operates between the feature extractor (\( \mathcal{F} \)) and the classifier (\( \mathcal{C} \)). In this PL without FM setting, modulation is applied when obtaining predictions for strongly augmented unlabeled data, weakly augmented labeled data, and during testing, but not when generating pseudo-labels from weakly augmented unlabeled data. In our method, we apply feature modulation during pseudo-label generation. The results show that incorporating modulation at this stage improves both the accuracy and utilization of pseudo-labels.

\section{Impact of uncertainty estimate}
We propose a new addition to the pseudo-label creation as the uncertainty estimate using MC dropout. We introduce a dropout layer to the ResNet-18 \cite{he2016deep} feature extractor after its last batch normalization layer with a dropout probability of 0.05. During the training process, we run five iterations of the complete model with feature extractor $f$,  feature modulator $\mathcal{M}$  and classifier $\mathcal{C}$ for the same data samples and compute the mean and standard deviation of the diagonal values of the prediction matrix. We use the calculated mean and standard deviation for pseudo-label generation as described in Sec. \ref{uncertanity_and_loss_scale}. Although this iterative process generates additional overhead of 2\%, it contributes to the overall accuracy improvement as described in Sec. \ref{sec:Experiments}. This modified pseudo-label mask is capable of improving the pseudo-label accuracy and improving the unlabaled data utilization. With the introduction of this uncertainty estimate, the pseudo-label threshold becomes adaptive based on the uncertainty of it. If the model is more certain about the class prediction, although the prediction probability is lower (around 0.75), we consider it as a correct label. On the other hand, if the uncertainty of the prediction is high, we disregard those labels despite their high prediction probability. As we introduce an adaptive threshold, we are able to reduce the fixed threshold down to
0.75.

\section{Hyperparameter Tuning}
We identify the learning rate of feature modulator, coefficients of labeled and unlabeled diagonal losses, $\beta$, $\gamma$ respectively as the hyperparameters. First, we selected suitable values for them as in Table \ref{tab:Loss scale version} considering the performance for OfficeHome dataset that has a lower variance. Then, we validated the performance of other datasets for the selected hyperparameters. We could gain consistent improvements for learning rate of 0.03, $\beta = 1.0$ and $\gamma = 0.5$.

\section{Performance compared to DG baselines}
In our method, we address the semi-supervised domain generalization problem. As mentioned in Sec.\ref{sec:intro} semi-supervised learning baselines perform better than domain generalization baselines. Here in Table \ref{DG} we compare the performance of SSDG problem setting of four common SSDG datasets when applied to DG baselines. 

\section{Limitations and Future Work}
We identify that our method has limitations when there are vast style distributions in the data, specifically in PACS dataset. We plan to mitigate this by increasing the robustness of our method to style shifts.

\end{document}